\documentclass[runningheads]{llncs}

\usepackage[year=2024,ID=8041]{eccv}

\usepackage{eccvabbrv}

\usepackage{graphicx}
\usepackage{booktabs}
\usepackage{url}
\usepackage{enumerate}
\usepackage{cancel}
\usepackage[accsupp]{axessibility}  

\usepackage[pagebackref,breaklinks,colorlinks,citecolor=eccvblue]{hyperref}

\usepackage{orcidlink}
\usepackage{xspace,mathtools,xcolor}
\newcommand{\vct}[1]{\boldsymbol{#1}\xspace}
\newcommand{\mat}[1]{\mathtt{#1}\xspace}
\newcommand{\aka}{\emph{a.k.a.}\xspace}

\begin{document}

\title{Revising Densification in Gaussian Splatting} 

\titlerunning{Revising Densification in Gaussian Splatting}

\author{Samuel Rota Bul\`o \and
Lorenzo Porzi \and
Peter Kontschieder}

\authorrunning{Rota Bul\`o et al.}

\institute{Meta Reality Labs Zurich\\
}

\maketitle

\begin{figure}
    \centering
    \includegraphics[width=\textwidth]{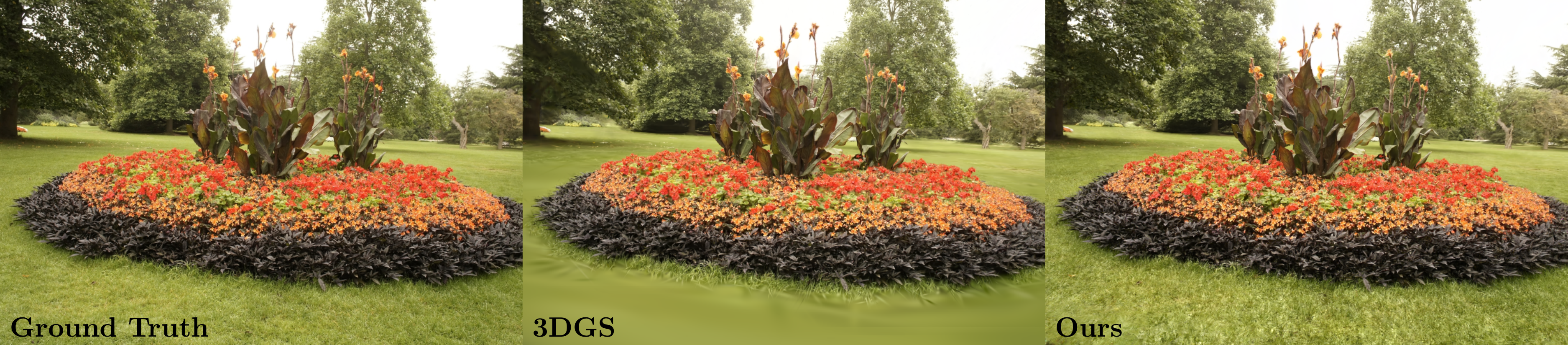}
    \caption{Densification is a critical component of 3D Gaussian Splatting (3DGS), and a common failure point. In this example (ground truth on the left) we show how 3DGS can fail (center) to add primitives to high-texture areas, like the grass in the bottom part of the pictures, producing large and blurry artifacts. Our approach (right) solves this issue by comprehensively revising densification in 3DGS.}
    \label{fig:enter-label}
    \vspace{-3em}
\end{figure}

\begin{abstract}

In this paper, we address the limitations of Adaptive Density Control (ADC) in 3D Gaussian Splatting (3DGS), a scene representation method achieving high-quality, photorealistic results for novel view synthesis. ADC has been introduced for automatic 3D point primitive management, controlling densification and pruning, however, with certain limitations in the densification logic. Our main contribution is a more principled, pixel-error driven formulation for density control in 3DGS, leveraging an auxiliary, per-pixel error function as the criterion for densification. We further introduce a mechanism to control the total number of primitives generated per scene and correct a bias in the current opacity handling strategy of ADC during cloning operations. Our approach leads to consistent quality improvements across a variety of benchmark scenes, without sacrificing the method's efficiency.

\end{abstract}

\section{Introduction}
\label{sec:intro}
High-quality, photorealistic scene modelling from images has been an important research area in computer vision and graphics, with plentiful applications in AR/VR/MR, robotics, \etc. In the last years, this field has gained a lot of attention due to advances in Neural 3D scene representations, particularly Neural Radiance Fields (NeRFs)~\cite{mildenhall2020NeRF}. NeRFs take a new approach to 3D scene representation and rendering, by leveraging a combination of deep learning and volumetric rendering techniques for generating photorealistic images from novel viewpoints. By optimizing MLPs to map from spatial coordinates and viewing directions to density and colour fields, these models have demonstrated astonishing capabilities for capturing the complex interplay of light and geometry in a data-driven way. While highly efficient in terms of representation quality, the original NeRF representation relies on time-consuming sampling strategies and thus excludes applications with fast rendering requirements. With many advances in terms of the underlying representation, these models have been significantly optimized towards improved training time and scene representation fidelity. However, inference speed for high-resolution, novel view synthesis remains an ongoing limitation. 

More recently, 3D Gaussian Splatting (3DGS)~\cite{kerbl3Dgaussians} has been proposed as an alternative and expressive scene representation, enabling both high-speed, high-fidelity training of models and high-resolution, GPU rasterization-friendly rendering of novel views. Their core representation is an optimized set of (anisotropic) 3D Gaussians, after being randomly distributed in 3D space, or systematically initialized at points obtained by Structure-from-Motion~\cite{snavely2006photo}. For obtaining a 2D image, all relevant 3D primitives are efficiently rendered via splatting-based rasterization with low-pass filtering. 

In 3DGS, each 3D primitive is parameterized as a 3D Gaussian distribution (\ie, with position and covariance), together with parameters controlling its opacity and describing its directional appearance (typically spherical harmonics). The parameter optimization procedure is guided by a multi-view, photometric loss, and is interleaved with Adaptive Density Control (ADC), a mechanism controlling density management for 3D points by means of introducing or deleting 3D primitives. ADC plays a critical role as it determines where to expand/shrink the scene representation budget for empty or over-reconstructed regions, respectively. Both growing and pruning operations are activated based on user-defined thresholds: Growing depends on the accumulated positional gradients of existing primitives and is, conditioned on the size of the Gaussians, executed by either splitting large primitives or by cloning smaller ones. Pruning is activated once the opacity falls below a provided threshold. While quite effective in practice, such density management strategies have several limitations. First, estimating a gradient magnitude-based threshold is rather non-intuitive and not robust to potential changes in the model, loss terms, \etc. Second, there are cases where only few and large Gaussians are modeling high-frequency patterns like grass as shown in the middle of~\cref{fig:enter-label}. Here, changes accumulated from positional gradients might remain very low and thus fail to trigger the densification mechanism, which in turn leads to substantial scene underfitting. Finally, ADC lacks explicit control of the maximum number of Gaussians generated per scene. This has important, practical implications as uncontrolled growth might easily lead to out-of-memory errors during training.

In this work we address the shortcomings of Adaptive Density Control proposed in the original 3D Gaussian splatting method. Our core contribution is a more principled, pixel-error driven formulation for density control in 3DGS. We describe how 2D, per-pixel errors as~\eg derived from Structural Similarity (or any other informative objective function) can be propagated back as errors to contributing Gaussian primitives. In our solution, we first break down the per-pixel errors according to each Gaussian's contribution, and in a camera-specific way. This allows us to track the maximum error per primitive for all views and across two subsequent ADC runs, yielding our novel, error-specific, and thus more intuitive decision criterion for densification. Our second contribution is correcting a bias introduced with the current form of opacity handling in ADC when conducting a primitive cloning operation. The original approach suggests to keep the same opacity for the cloned Gaussian, which however biases the alpha-compositing logic applied for rendering the pixel colors. Indeed, this procedure leads to an overall increase of opacity in the cloned region, preventing the model to correctly account for contributions of other primitives and thus negatively affecting the densification process.
Our third contribution is a mechanism for controlling the total number of primitives generated per scene and the maximum amount of novel primitives introduced per densification run. With this functionality, we can avoid undesired out-of-memory errors and better tune the method's behaviour~\wrt given hardware constraints.
We extensively validate our contributions on standard benchmark datasets like Mip-NeRF 360~\cite{barron2022mip}, Tanks and Temples~\cite{knapitsch2017tanks}, and Deep Blending~\cite{hedman2018deep}. Our experiments show consistent improvements over different baselines including 3DGS~\cite{kerbl3Dgaussians} and Mip-Splatting~\cite{yu2023mip}.
To summarize, our contributions are improving methodological shortcomings in 3DGS'~Adaptive Density Control mechanism as follows:
\begin{itemize}
    \item We propose a principled approach that enables the guidance of the densification process according to an auxiliary, per-pixel error function, rather than relying on positional gradients.
    \item We correct an existing, systematic bias from the primitive growing procedure when cloning Gaussians, negatively impacting the overall densification.
    \item We present ablations and experimental evaluations on different, real-world benchmarks, confirming quantitative and qualitative improvements. 
\end{itemize}

\subsection{Related works}
Since it was presented in~\cite{kerbl3Dgaussians}, 3DGS has been used in a remarkably wide set of downstream applications, including Simultaneous Localization and Mapping~\cite{matsuki2023gaussian,yugay2023gaussian,yan2023gs,keetha2023splatam}, text-to-3D generation~\cite{chen2023text,tang2023dreamgaussian,yi2023gaussiandreamer}, photo-realistic human avatars~\cite{zielonka2023drivable,lei2023gart,kocabas2023hugs,saito2023relightable}, dynamic scene modeling~\cite{wu20234d,luiten2023dynamic,yang2023real} and more~\cite{guedon2023sugar,xie2023physgaussian,ye2023gaussian}.
However, only a handful of works like ours have focused on advancing 3DGS itself, by improving its quality or overcoming some of its limitations.

In GS++~\cite{huang2024gs++}, Huang~\etal present an improved approximation of the 3D-to-2D splatting operation at the core of 3DGS, which achieves better accuracy near image edges and solves some common visual artifacts.
Spec-Gaussian~\cite{yang2024spec} and Scaffold-gs~\cite{lu2023scaffold} focus on improving view-dependent appearance modeling: the former by replacing spherical harmonics with an anisotropic spherical Gaussian appearance field; the latter by making all 3D Gaussian parameters, including whether specific primitives should be rendered or not, dependent on view direction through a small MLP.
Mip-Splatting~\cite{yu2023mip} tackles the strong artifacts that appear in 3DGS models when they are rendered at widely different resolutions (or viewing distances) compared to the images they were trained on.
To do this, Yu~\etal propose to incorporate a 3D filter to constrain the size of the 3D primitives depending on their maximal sampling rate on the training views, and a 2D Mip filter to mitigate aliasing issues.
All these works adopt the original ADC strategy proposed in~\cite{kerbl3Dgaussians}, and can potentially benefit from our improved approach, as we show for Mip-Splatting in Sec.~\ref{sec:experiments}.

Only few concurrent works have touched on densification, while putting most of their focus on other aspects of 3DGS.
Lee~\etal~\cite{lee2023compact} propose a quantization-based approach to produce more compact 3DGS representations, which includes a continuous sparsification strategy that takes both primitive size and opacity into account.
GaussianPro~\cite{cheng2024gaussianpro} directly tackles related densification limitations as we explore in our work, filling the gaps from SfM-based initialization.
They propose a rather complex procedure based on the progressive propagation of primitives along estimated planes, using patch-matching and geometric consistency as guidance.
In contrast to our method,~\cite{cheng2024gaussianpro} focuses on fixing the quality of planar regions, instead of holistically improving densification.
We also note that a fair comparison with their method on the standard Mip-NeRF 360 benchmark is not feasible at the time of submission, as the authors did not publicly share the improved SfM point cloud used in their experiments (see \S 5.2 of~\cite{cheng2024gaussianpro}).

\section{Preliminaries: Gaussian Splatting}
\label{sec:prelim}

Gaussian Splatting~\cite{kerbl3Dgaussians} revisits ideas from EWA splatting~\cite{zwickerVolSplatting} and proposes to fit a 3D scene as a collection of 3D Gaussian primitives $\Gamma\coloneqq\{\gamma_1,\ldots,\gamma_K\}$ that can be rendered by leveraging volume splatting.

\paragraph{Gaussian primitive.}
A Gaussian primitive $\gamma_k\coloneqq(\vct\mu_k, \mat\Sigma_k,\alpha_k, \vct f_k)$ geometrically resembles a  3D Gaussian kernel
\[
\mathcal G_k(\vct x)\coloneqq \exp\left(-\frac{1}{2}(\vct x-\vct\mu_k)^\top\mat\Sigma_k^{-1}(\vct x-\vct\mu_k)\right)
\]
centered in $\vct\mu_k\in\mathbb R^3$ and having $\mat\Sigma_k$ as its $3\times 3$ covariance matrix. Each primitive additionally entails 
an opacity factor $\alpha_k\in[0,1]$ and 
a feature vector $\vct f_k\in\mathbb R^d$ (\eg~RGB color or spherical harmonics coefficients). 

\paragraph{Splatting.} 
This is the operation of projecting a Gaussian primitive $\gamma_k$ to a camera pixel space via its world-to-image transformation $\pi:\mathbb R^3\to\mathbb R^2$, which we refer directly 
 to as the camera for simplicity. The projection $\pi$ is approximated to the first order at the primitive's center $\vct\mu_k$ so that the projected primitive is geometrically equivalent to a 2D Gaussian kernel $\mathcal G_k^\pi$ with mean $\pi(\vct\mu_k)\in\mathbb R^2$ and 2D covariance $\mat J_k^{\pi}\mat\Sigma_k \mat {J_k^{\pi}}^\top$ with $\mat J_k^{\pi}$ being the Jacobian of $\pi$ evaluated at $\vct\mu_k$.

\paragraph{Rendering.} 
To render the primitives $\Gamma$ representing a scene from camera $\pi$, we require a decoder $\Phi$ to be specified, which provides the feature we want to render as $\Phi(\gamma_k,\vct u)\in\mathbb R^m$ for each Gaussian primitive $\gamma_k$ and pixel $\vct u$. Moreover, we assume Gaussian primitives $\Gamma$ to be ordered with respect to their center's depth, when seen from the camera's reference frame.
Then, the rendering equation takes the following form (with $\Gamma$ being omitted from the notation)
\[
\mathcal R[\pi,\Phi](\vct u)\coloneqq {\sum_{k=1}^{K}}\Phi(\gamma_k,\vct u)\omega^\pi_k(\vct u)\,,
\]
where $\omega^\pi_k(\vct u)$ are alpha-compositing coefficients given by
\[
\omega^\pi_k(\vct u)\coloneqq\alpha_k\mathcal G^\pi_k(\vct u)\prod_{j=1}^{k-1}\left(1-\alpha_j\mathcal G^\pi_j(\vct u)\right).
\]
If we assume the feature vectors $\vct f_k$ to be spherical harmonics coefficients encoding an RGB function on the sphere, we can regard $\Phi_{\mathtt{RGB}}(\vct u)$ as the 
decoded RGB color for the given view direction associated to pixel $\vct u$.
If we use  $\Phi_\mathtt {RGB}$ as the decoder in the rendering equation, we obtain a rendered color image $C_\pi(\vct u)\coloneqq\mathcal R[\pi,\Phi_{\mathtt {RGB}}](\vct u)$ for each camera $\pi$.
Similarly, one can pick different $\Phi$'s to enable the rendering of depth, normals, or other quantities of interest as we will show later.

\paragraph{Mip-splatting.} In~\cite{yu2023mip}, the authors introduce a variation of standard Gaussian splatting that focuses on solving aliasing issues. We refer the reader to the original paper for details, but the idea is to track the maximum sampling rate for each Gaussian primitive and use it to reduce aliasing effects by attenuating the Gaussian primitives' opacity.

\section{Revising Densification}
We first review the Adaptive Density Control module proposed in the original Gaussian splatting work~\cite{kerbl3Dgaussians}, highlight some of its limitations, and then introduce our novel and improved densification procedure.

\subsection{Adaptive Density Control and its limitations}
\label{sec:adc_limitations}

3DGS~\cite{kerbl3Dgaussians} and follow-up extensions (e.g. Mip-splatting~\cite{yu2023mip}) rely on the Adaptive Density Control (ADC) module to grow or prune Gaussian primitives. This module is run according to a predetermined schedule and densification decisions are based on gradient statistics collected across the ADC runs.
Specifically, for each Gaussian primitive $\gamma_k$ the positional gradient magnitude $\left\Vert\frac{\partial L_\pi}{\partial\vct \mu_k}\right\Vert$ is tracked and averaged over all rendered views $\pi\in\Pi$ within the collection period, where $L_\pi$ denotes the loss that is optimized for camera $\pi$. The resulting quantity is denoted by $\tau_k$.

\paragraph{Growing.} ADC grows new Gaussian primitives via a \emph{clone} or a \emph{split} operation. A primitive $\gamma_k$ will be considered for a growing operation only if $\tau_k$ exceeds a user-defined threshold. The decision about which operation to apply depends on the size of the primitive measured in terms of the largest eigenvalue of the covariance matrix $\mat\Sigma_k$. Specifically, primitives larger than a threshold are split, otherwise cloned. When a primitive $\gamma_k$ is split, two new primitives are generated with their position being sampled from $\mathcal{G}_k$ and their covariance being a scaled down version of $\mat\Sigma_k$, while preserving the same opacity and feature vector. When a clone operation takes place, a simple clone of $\gamma_k$ is instantiated.

\paragraph{Pruning.} ADC prunes a Gaussian primitive $\gamma_k$ if its opacity $\alpha_k$ is below a user-defined threshold, typically $0.005$. To ensure that an unused primitive is eventually pruned, a hard-reset of the opacity to a minimum value (usually $0.01$) is enforced according to a predefined schedule.

\paragraph{Limitations.} Deciding which Gaussian primitives to split/clone based on the magnitude of the positional gradient suffers from a number of limitations: 
\begin{itemize}
\item Determining a threshold for a gradient magnitude is not intuitive and very sensitive to modifications to the model, losses and hyperparameters,
\item There are cases of scene underfitting also when the value of $\tau_k$ is below the threshold that triggers densification (see \cref{fig:enter-label}).
\item It is not possible to directly control the number of Gaussian primitives that are generated for a given scene, resulting in possible out-of-memory errors if their number grows abnormally.
\end{itemize}
In addition, we found that the ADC's logic of growing primitives suffers from a bias that weights more the contribution of freshly cloned primitives. More details will follow in \cref{sec:opacity_correction}

\subsection{Error-based densification}
\label{sec:densification}

Assume we have an image with an area characterized by a high-frequency pattern and covered by few large splatted Gaussian primitives (\eg the grass in \cref{fig:enter-label}). Under this scenario, an infinitesimal change in the 3D location $\vct\mu_k$ of one of the corresponding Gaussian primitives $\gamma_k$ will leave the error almost unchanged and, hence, the collected magnitude of the positional gradient $\tau_k$ remains close to zero. In fact, $\tau_k$ is sensitive to error-changes, but is blind to the absolute value of the error. This becomes a problem, for we expect to increase the number of Gaussian primitives in areas exhibiting a larger error. 

Given the above considerations, we propose to steer the densification decisions directly based on an auxiliary per-pixel error function $\mathcal E_\pi$ (\eg~Structural Similarity) that we measure when rendering on a camera $\pi$ with available ground-truth.
One problem to address is how to turn per-pixel errors into per-Gaussian-primitive errors in light of the fact that each pixel error entangles the contribution of multiple Gaussian primitives. Our solution consists of first re-distributing the per-pixel errors $\mathcal E_\pi(\vct u)$ to each Gaussian primitive $\gamma_k$ proportionally to their contribution to the rendered pixel color, \ie proportionally to $w^\pi_k(\vct u)$. This yields the following error for each primitive $\gamma_k$ and camera $\pi$:
\[
E^\pi_k\coloneqq \sum_{\vct{u}\in\text{Pix}} \mathcal E_\pi(\vct u)w^\pi_k(\vct u)\,,
\]
where the sum runs over the image pixels.
Then, for each primitive $\gamma_k$ we track the maximum value of the error $E^\pi_k$ across all views $\pi\in\Pi$ seen between two runs of the ADC module, \ie 
\[
E_k\coloneqq \max_{\pi\in\Pi} E^\pi_k.
\]
This is the score that we use to prioritize the growing of Gaussian primitives. 
As opposed to $\tau_k$, it is easier to set a threshold for our new densification score, for it is typically expressed in terms of a known error metric.

\paragraph{Implementation details.} In order to compute $E^\pi_k$ we assign an additional scalar $e_k$ to each Gaussian primitive $\gamma_k$, and enable the possibility of rendering it via the decoder $\Phi_\mathtt{ERR}(\gamma_k,\vct u)\coloneqq e_k$. Then, we add the following auxiliary loss to the standard Gaussian splatting training objective:
\[
L^\mathtt{aux}_\pi\coloneqq \sum_{\vct u\in\text{Pix}} \cancel\nabla[\mathcal E_\pi(\vct u)]\underbrace{\mathcal R[\pi, \Phi_\mathtt{ERR}](\vct u)}_{=\sum_{k=1}^K e_k\omega^\pi_k(\vct u)}\,,
\]
which is basically the dot product of the per-pixel error with gradient detached and the rendering of the newly-added scalar.
We initialize $e_k$ to $0$ for each Gaussian primitive $\gamma_k$ and never update it during training. In this way, $L^\mathtt{aux}_\pi=0$ and all Gaussian primitives' parameters, excepting $e_k$, are left invariant by this loss. The gradient with respect to $e_k$ instead yields
\[
\frac{\partial L^\mathtt{aux}_\pi}{\partial e_k}=\sum_{\vct{u}\in\text{Pix}} \mathcal E_\pi(\vct u)\omega^\pi_k(\vct u)=E^\pi_k\,,
\]
which is the per-Gaussian-primitive error for camera $\pi$ we wanted to compute. 

\subsection{Opacity correction after cloning}\label{sec:opacity_correction}
\begin{figure}[t]
    \centering
    \includegraphics[width=\linewidth]{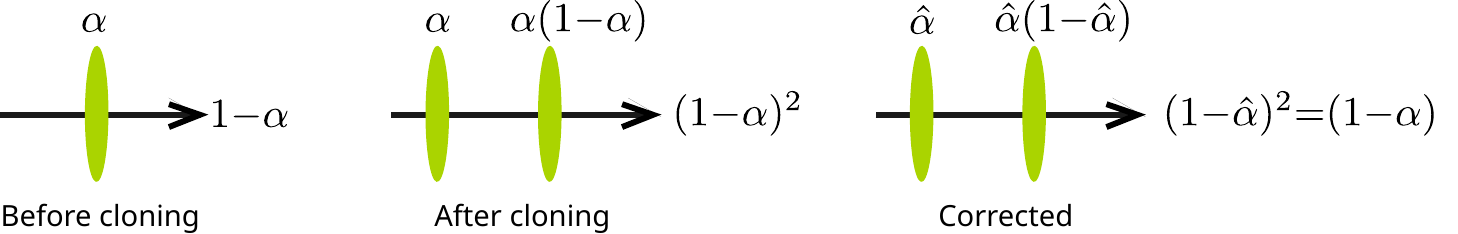}
    \caption{Consider rendering a single splatted Gaussian in its center pixel with opacity $\alpha$ before and after cloning. Before we clone, the rendered color depends with weight $1-\alpha$ on what comes next. After we clone, since we preserve the opacity, the rendered color depends with weight $(1-\alpha)^2$ on what comes next. Since $(1-\alpha)\geq(1-\alpha)^2$ we have a bias towards weighting more Gaussian primitives that get cloned. The proposed correction changes the opacity post clone to $\hat\alpha$ so that the bias is removed.}
    \label{fig:opacity_correction}
    \vspace{-10pt}
\end{figure}

In the original ADC module, when a Gaussian primitive is split or cloned, the opacity value is preserved. This choice introduces a bias in the case of the clone operation by implicitly increasing the impact of the densified primitive on the final rendered color. To see why this is the case we can follow the example in \cref{fig:opacity_correction}, where we consider what happens if we render a splatted Gaussian in its center pixel assuming an opacity value $\alpha$. Before a cloning operation happens, the rendered color depends on primitives that come next in the ordering  with weight $1-\alpha$. But after we clone, due to the alpha-compositing logic, we have that primitives that come next weight $(1-\alpha)^2$, which is lower than $1-\alpha$ for all opacity values in $(0,1)$. Accordingly, by applying the standard logic of preserving the opacity after cloning we have a bias to weight more the cloned primitives. 
The solution we suggest consist in reducing the opacity of the primitives after cloning so that the bias is removed. The new opacity value $\hat\alpha$ can be found by solving the equation $(1-\alpha)=(1-\hat\alpha)^2$, which yields $\hat\alpha\coloneqq1-\sqrt{1-\alpha}$.

If we depart from the simplified setting of considering only the center pixel and rather consider all pixels, it is unfortunately not possible to remove completely the bias.  Nonetheless, the correction factor we introduce reduces the bias for \emph{all} pixels compared to keeping the opacity of the cloned primitive. Indeed, the following relation holds for all $\alpha_k\in(0,1)$ and all pixels $\vct u$:
\[
1-\alpha_k\mathcal G_k^\pi(\vct u)\geq(1-\hat\alpha_k\mathcal G_k^\pi(\vct u))^2>(1-\alpha_k\mathcal G_k^\pi(\vct u))^2\,,
\]
where $\hat\alpha_k\coloneqq 1-\sqrt{1-\alpha_k}$ is our corrected opacity. The proof of the relation follows by noting that $\hat\alpha_k$ can be rewritten as $\frac{\alpha_k}{1+\sqrt{1-\alpha_k}}$ which is strictly smaller than $\alpha_k$ for $\alpha_k\in(0,1)$.

Finally, the correction of the opacity as shown above is derived assuming we clone a Gaussian primitive, but does not strictly match the case of a split operation, for when we split we move the two offspring randomly away from the previous center and we change the covariance scale. For this reason, we stick to the standard rule of preserving the opacity of a primitive we split.

\subsection{Primitives growth control}\label{sec:growth}

\begin{figure}[t]
    \centering
    \includegraphics[width=0.7\textwidth]{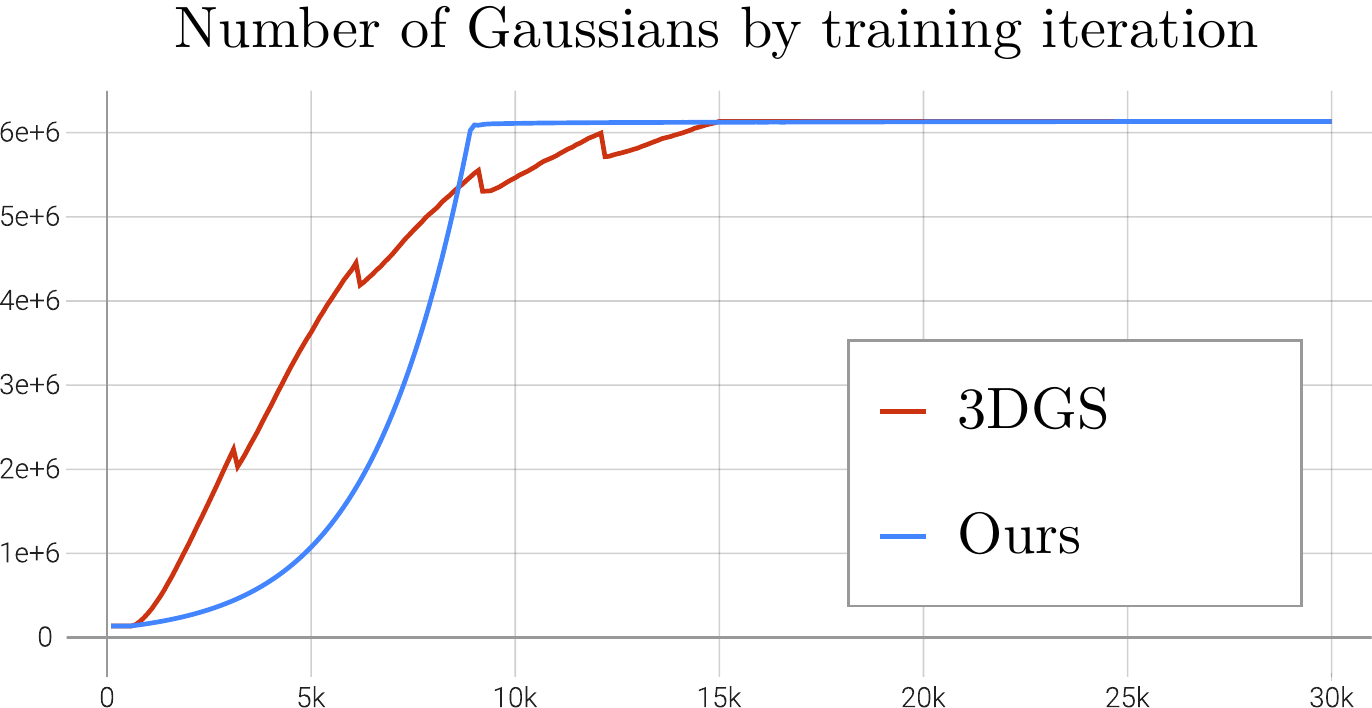}
    \caption{Evolution of the number of Gaussians in 3DGS, and in our method with upper limit set to the number reached by 3DGS (on the \texttt{garden} scene from the Mip-NeRF 360 dataset). Note that, while 3DGS' ADC process stops after 15k iterations, ours remains active for 27k. This is not immediately visible from the plot, since pruned primitives are immediately replaced by newly spawned ones, keeping the overall number stable once the maximum is reached.}
    \label{fig:growth}
\end{figure}

The ADC module grows a Gaussian primitive if $\tau_k$ is larger than a threshold.
This mechanism can lead to unpredictable growth of the number of primitives, eventually resulting in out-of-memory issues.
To avoid this problem, we introduce a global limit to the maximum number of Gaussian primitives, and a mechanism to control the maximum number of primitives that can be created each time densification is run.
Among the many possible options, we explore a logic that limits new primitive offspring to a fixed fraction of the primitives that already exist.
In case the number of primitives that are entitled to be densified exceeds the available budget, we retain only the ones that exhibit the highest densification score.
An example of this process is shown in Fig.~\ref{fig:growth}, compared to the one from 3DGS: For \textit{Ours}, the number of primitives grows smoothly until it reaches the allotted maximum, without the discontinuities induced by opacity reset (see Sec.~\ref{sec:opacity}).
The way we control the number of primitives is not limited to our error-based densification logic, but can be applied equally to the original gradient-based one.

\subsection{Alternative to opacity reset}\label{sec:opacity}
The strategy introduced in~\cite{kerbl3Dgaussians} to favour the sparsification of Gaussian primitives consists in periodical hard resets of the opacity for all primitives to a low value, so that primitives whose opacity is not increased again by the optimization will eventually be pruned.
This introduces a small shock in the training trajectory, which is suboptimal for the sake of having stable and predictable training dynamics.
Moreover, resetting the opacity is particularly harmful for our error-based densification method, for it will lead to misleading error statistics right after the hard-reset, potentially triggering wrong densification decisions.
For this reason, we propose a different logic to favour primitives pruning in a smoother way.
Specifically, we decrease the opacity of each primitive by a fixed amount (we use $0.001$) after each densification run, so that the opacity will gradually move towards the pruning range.
In this way, we avoid sudden changes in the densification metric, while preserving the desired sparsification properties. 

One downside of the new opacity regularization logic is that the constant push towards lowering the opacity of the primitives implicitly invites the model to make more use of the background where possible.
This is also harmful, for it could generate more holes in the scene that will be visible from novel views.
To counteract this dynamics, we also regularize the residual probabilities of the alpha-compositing (\aka residual transmittance) to be zero for every pixel, by simply minimizing their average value, weighted by a hyperparameter (here $0.1$).

\section{Experimental Evaluation}
\label{sec:experiments}

In the following we show how our improved ADC mechanism can equally be applied both to standard 3DGS~\cite{kerbl3Dgaussians} and its Mip-Splatting extension~\cite{yu2023mip}, providing benefits to both.

\subsection{Datasets and metrics}

We follow the experimental setup from the 3DGS~\cite{kerbl3Dgaussians} paper, focusing on the real-world scenes from the Mip-NeRF 360~\cite{barron2022mip}, Tanks and Temples~\cite{knapitsch2017tanks} and Deep Blending~\cite{hedman2018deep} datasets.
Mip-NeRF 360 comprises nine scenes (5 outdoor, 4 indoor) captured in a circular pattern which focuses on a central area of a few meters, with a potentially unbounded background.
For Tanks and Temples, we focus on the ``Truck'' and ``Train'' scenes, while for Deep Blending we focus on the ``Dr Johnson'' and ``Playroom'' scenes, using the images and SfM reconstructions shared by the Gaussian Splatting authors.
In each experiment we set aside each 8th image as a validation set, and report peak signal-to-noise ratio (PSNR), structural similarity (SSIM) and the perceptual metric from~\cite{zhang2018unreasonable} (LPIPS).

\subsection{Experimental setup}
\label{sec:exp_setup}

We evaluate based on our re-implementations of 3DGS, which allows us to easily switch between standard 3DGS, Mip-Splatting, the original ADC of~\cite{kerbl3Dgaussians}, our contributions or any combination thereof.
We reproduce the training settings proposed in~\cite{yu2023mip,kerbl3Dgaussians} and the respective public code-bases\footnote{\url{https://github.com/graphdeco-inria/gaussian-splatting}}\footnote{\url{https://github.com/autonomousvision/mip-splatting}}, including number of training iterations, batch size, input resolution, learning rates etc.
When training with our contributions, we grow Gaussians with $E_k > 0.1$, adding up to $5\%$ of the current number of primitives at each densification step.
Differently from 3DGS, we keep our ADC process active for 27k iterations (\ie 90\% of the training process), instead of stopping it after 15k.
Other relevant hyper-parameters are left to the default values used in 3DGS, and shared across all datasets and scenes.
In all our experiments, we set the maximum primitives budget to the number of primitives (or its median, for experiments with multiple runs) generated by the corresponding baseline, in order to obtain perfectly comparable models.
For more details, please refer to the supplementary document.

\paragraph{A note about LPIPS.} Investigating the 3DGS and Mip-Splatting baselines, we discovered a discrepancy in the way LPIPS is calculated in both public code-bases, which resulted in under-estimated values being reported in the original papers.
This was confirmed in private correspondence with the authors.
In order to simplify comparisons with future works that don't rely on these code-bases, and might be unaware of this issue, we report \emph{correct LPIPS values} here, and refer the reader to the supplementary document for values compatible with those shown in the tables of~\cite{yu2023mip,kerbl3Dgaussians}.

\subsection{Main results}

\begin{figure}
    \centering
    \includegraphics[width=\textwidth]{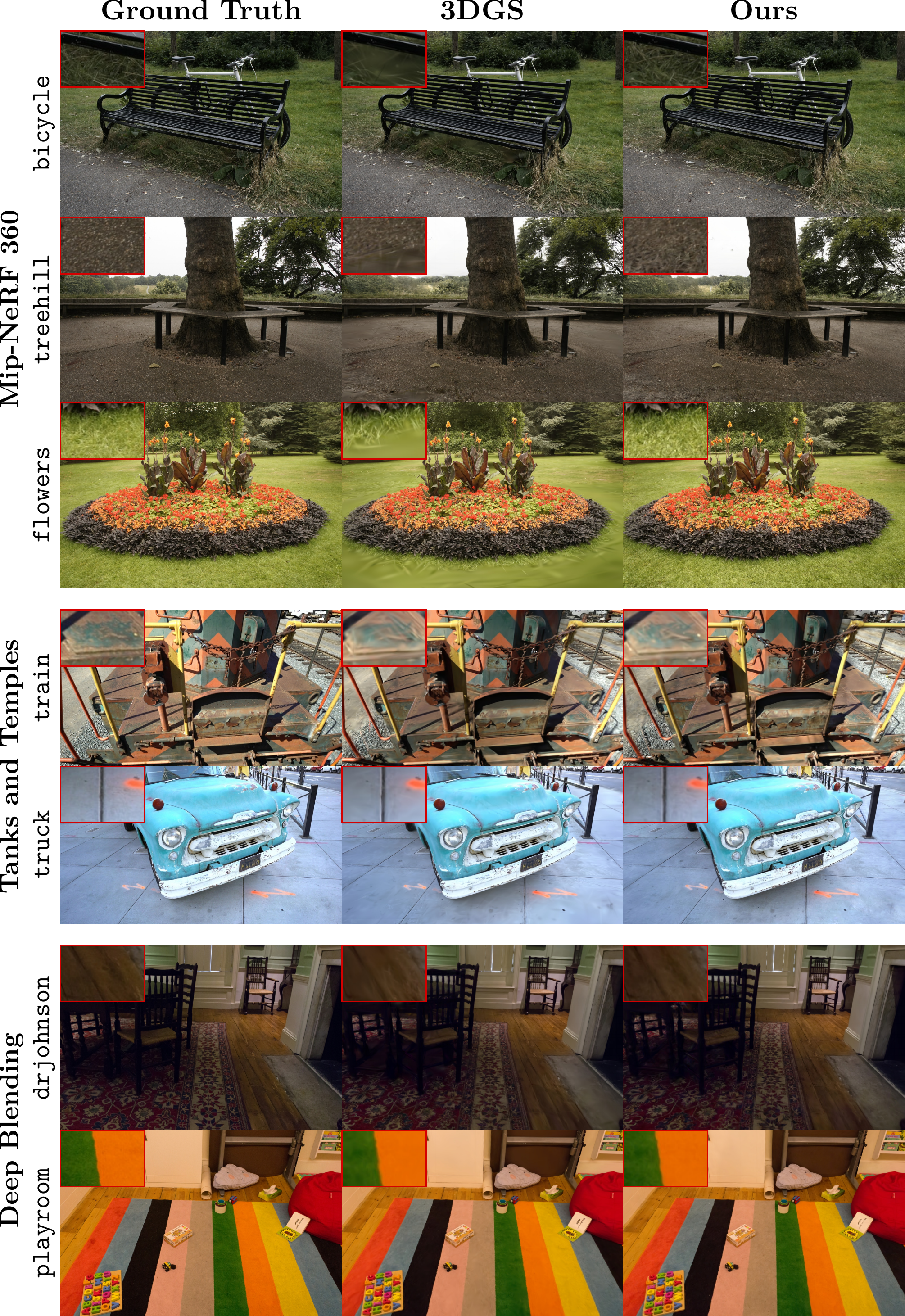}
    \caption{Qualitative results on the Mip-NeRF 360, Tanks and Temples and Deep Blending validation sets.
    Note that 3DGS and Ours use \emph{the same number of primitives}.
    Best viewed on screen at high magnification.}
    \label{fig:qual_results}
\end{figure}

\begin{table}[t]
    \centering
    \caption{Results on the Mip-NeRF 360 dataset. Top section of the table: results from the Gaussian Splatting paper; bottom section: results from our re-implementation averaged over 5 runs.}
    \label{tab:results_mipnerf360}
    \begin{tabular}{l|cc|cc|cc}
    \toprule
         & \multicolumn{2}{c|}{PSNR $\uparrow$} & \multicolumn{2}{c|}{SSIM $\uparrow$} & \multicolumn{2}{c}{LPIPS $\downarrow$} \\
        Method & Mean & Stdev & Mean & Stdev & Mean & Stdev \\
    \midrule
        Plenoxels~\cite{fridovich2022plenoxels} & 23.08 & -- & 0.626 & -- & 0.436 & -- \\
        INGP~\cite{mueller2022instant} & 25.59 & -- & 0.699 & -- & 0.331 & -- \\
        Mip-NeRF 360~\cite{barron2022mip} & 27.69 & -- & 0.792 & -- & 0.237 & -- \\
    \midrule
        3DGS~\cite{kerbl3Dgaussians} & 27.45 & 0.081 & 0.817 & 0.001 & 0.250 & 0.001 \\
        Mip-Splatting~\cite{yu2023mip} & 27.51 & 0.074 & 0.818 & 0.001 & 0.251 & 0.001 \\
        Ours, 3DGS & 27.61 & 0.067 & 0.822 & 0.001 & \textbf{0.223} & 0.001 \\
        Ours, Mip-Splatting & \textbf{27.70} & 0.065 & \textbf{0.823} & 0.001 & \textbf{0.223} & 0.001 \\
    \bottomrule
    \end{tabular}
\end{table}

\begin{table}[t]
    \centering
    \caption{Results on the Tanks and Temples dataset. Top section of the table: results from the Gaussian Splatting paper; bottom section: results from our re-implementation averaged over 5 runs.}
    \label{tab:results_tandt}
    \begin{tabular}{l|cc|cc|cc}
    \toprule
         & \multicolumn{2}{c|}{PSNR $\uparrow$} & \multicolumn{2}{c|}{SSIM $\uparrow$} & \multicolumn{2}{c}{LPIPS $\downarrow$} \\
        Method & Mean & Stdev & Mean & Stdev & Mean & Stdev \\
    \midrule
        Plenoxels~\cite{fridovich2022plenoxels} & 21.08 & -- & 0.719 & -- & 0.379 & -- \\
        INGP~\cite{mueller2022instant} & 21.92 & -- & 0.745 & -- & 0.305 & -- \\
        Mip-NeRF 360~\cite{barron2022mip} & 22.22 & -- & 0.759 & -- & 0.257 & -- \\
    \midrule
        3DGS~\cite{kerbl3Dgaussians} & 23.44 & 0.089 & 0.845 & 0.002 & 0.212 & 0.001 \\
        Mip-Splatting~\cite{yu2023mip} & 23.65 & 0.066 & 0.849 & 0.001 & 0.211 & 0.001 \\
        Ours, 3DGS & 23.93 & 0.070 & 0.853 & 0.001 & 0.187 & 0.001 \\
        Ours, Mip-Splatting & \textbf{24.10} & 0.109 & \textbf{0.857} & 0.002 & \textbf{0.183} & 0.001 \\
    \bottomrule
    \end{tabular}
\end{table}

\begin{table}[t]
    \centering
    \caption{Results on the Deep Blending dataset. Top section of the table: results from the Gaussian Splatting paper; bottom section: results from our re-implementation averaged over 5 runs.}
    \label{tab:results_db}
    \begin{tabular}{l|cc|cc|cc}
    \toprule
         & \multicolumn{2}{c|}{PSNR $\uparrow$} & \multicolumn{2}{c|}{SSIM $\uparrow$} & \multicolumn{2}{c}{LPIPS $\downarrow$} \\
        Method & Mean & Stdev & Mean & Stdev & Mean & Stdev \\
    \midrule
        Plenoxels~\cite{fridovich2022plenoxels} & 23.06 & -- & 0.795 & -- & 0.510 & -- \\
        INGP~\cite{mueller2022instant} & 24.96 & -- & 0.817 & -- & 0.390 & -- \\
        Mip-NeRF 360~\cite{barron2022mip} & 29.40 & -- & 0.901 & -- & \textbf{0.245} & -- \\
    \midrule
        3DGS~\cite{kerbl3Dgaussians} & 29.54 & 0.096 & 0.902 & 0.000 & 0.311 & 0.001 \\
        Mip-Splatting~\cite{yu2023mip} & \textbf{29.68} & 0.068 & 0.903 & 0.000 & 0.309 & 0.000 \\
        Ours, 3DGS & 29.50 & 0.110 & 0.904 & 0.001 & 0.305 & 0.001 \\
        Ours, Mip-Splatting & 29.64 & 0.126 & \textbf{0.905} & 0.001 & 0.303 & 0.001 \\
    \bottomrule
    \end{tabular}
\end{table}

In a first set of experiments, we evaluate the effectiveness of our improved ADC strategy (Ours) when applied to 3DGS and Mip-Splatting.
Results, collected over 5 training runs to average out the randomness induced by stochastic primitive splitting, are reported in Tab.~\ref{tab:results_mipnerf360}, \ref{tab:results_tandt} and~\ref{tab:results_db}.
For the sake of completeness, we also include scores obtained with three NeRF baselines, \ie Plenoxels~\cite{fridovich2022plenoxels}, Instant-NGP (INGP)~\cite{mueller2022instant} and Mip-NeRF 360~\cite{barron2022mip}, as originally reported in~\cite{kerbl3Dgaussians}.
Our approach consistently outperforms the corresponding baselines (\ie Ours, 3DGS vs. 3DGS; Ours, Mip-Splatting vs. Mip-Splatting), particularly on SSIM and LPIPS.
This is in line with what we discussed in Sec.~\ref{sec:adc_limitations} and~\ref{sec:densification}: standard ADC often leads to localized under-fitting, as it fails to split large gaussians that cover highly-textured regions of the scene.
This kind of errors is poorly reflected by PSNR, which measures the ``average fit'' over image pixels, but are promptly detected by perceptual metrics like LPIPS.
On the Deep Blending dataset, we observe thinner gaps, PSNR actually showing a small regression \wrt the baselines, although with low confidence margins.
We suspect this might be related to the fact that Deep Blending contains many flat, untextured surfaces (see Fig.~\ref{fig:qual_results}), that are particularly challenging to reconstruct accurately with 3DGS-like methods, independently of the ADC strategy being adopted.

Figure~\ref{fig:qual_results} contains a qualitative comparison between standard 3DGS and 3DGS augmented with our contributions (Ours).
Areas with under-fitting artifacts are highlighted, showing how these are notably ameliorated by our approach.
It is also worth noting that Ours effectively maintains the same quality as 3DGS in non-problematic areas, producing a more perceptually accurate reconstruction while using \emph{the same number of primitives} (see Sec.~\ref{sec:exp_setup}).

\subsection{Ablation experiments}

\begin{table}[t]
    \centering
    \caption{Ablation experiments on the Mip-NeRF 360 dataset, adding individual contributions to 3DGS or removing them from Ours. OC: Opacity Correction, Sec.~\ref{sec:opacity_correction}; GC: Growth Control, Sec.~\ref{sec:growth}; OR: Opacity Regularization, Sec.~\ref{sec:opacity}.}
    \label{tab:ablations}
    \begin{tabular}{l|cccc|cccc}
    \toprule
    & \multicolumn{4}{c|}{3DGS} & \multicolumn{4}{c}{Ours} \\
    & Baseline & $+$OC & $+$GC & $+$OR & Full & $-$OC & $-$GC & $-$OR \\
    \midrule
    PSNR $\uparrow$
        & 27.45 & 27.65 & 27.35 & 27.48
        & 27.61 & 27.04 & 27.54 & 27.28 \\
    SSIM $\uparrow$
        & 0.817 & 0.822 & 0.810 & 0.818
        & 0.822 & 0.812 & 0.818 & 0.810 \\
    LPIPS $\downarrow$
        & 0.250 & 0.239 & 0.256 & 0.243
        & 0.223 & 0.235 & 0.223 & 0.234 \\
    \bottomrule
    \end{tabular}
    \vspace{-10pt}
\end{table}

In Tab.~\ref{tab:ablations} we ablate the effects of Opacity Correction (OC, Sec.~\ref{sec:opacity_correction}), Growth Control (GC, Sec.~\ref{sec:growth}) and Opacity Regularization (OR, Sec.~\ref{sec:opacity}) on the Mip-NeRF 360 dataset.
In particular, we evaluate 3DGS augmented with each of these components (left side of the table), and our method with the components replaced by the corresponding baseline mechanism in 3DGS' standard ADC (right side of the table).
First, we observe that OC, GC and OR all contribute to our method, as the Full version of Ours achieves the overall best results on all metrics, and removing them consistently degrades performance.
Interestingly, Opacity Correction seems to have the largest impact here, as it produces both the largest increase in the scores when added to 3DGS, and the largest decrease when removed from Ours.
Finally, Growth Control has a negative impact on 3DGS when utilized in isolation, while only slightly degrading the results when removed from Ours.
Note that this observation doesn't detract from GC's usefulness as a strategy to control and limit the capacity of the model.
We hypothesize that GC's negative effect on 3DGS might be a consequence of the fact that the standard, gradient-based densification score is actually a poor choice for comparing gaussians in terms of how soon they should be split or cloned (remember that GC ranks Gaussians based on their score).

\subsection{Limitations}

While our method appears to be quite effective at solving under-fitting issues, these can still be present in especially difficult scenes (\eg \texttt{treehill} in the Mip-NeRF 360 dataset, both scenes from the Deep Blending dataset).
Focusing on the problematic areas that our ADC approach handles successfully, we observe that, while perceptually more ``correct'', the reconstruction there can still be quite inaccurate when closely compared to the ground truth (see \eg the \texttt{flowers} scene in Fig.~\ref{fig:qual_results}).
We suspect both these issues might be related to 3DGS' intrinsic limits in handling i) strong view-dependent effects; ii) appearance variations across images; and iii) errors induced by the linear approximation in the Splatting operation (see Sec.~\ref{sec:prelim}).
An interesting future direction could be to combine our approach with works that address these issues, \eg Spec-Gaussian~\cite{yang2024spec} for (i) and GS++~\cite{huang2024gs++} for (iii).

\section{Conclusion}
In this paper, we addressed the limitations of the Adaptive Density Control (ADC) mechanism in 3D Gaussian Splatting (3DGS), a scene representation method for high-quality, photorealistic rendering. Our main contribution is a more principled, pixel-error driven formulation for density control in 3DGS. We propose how to leverage a novel decision criterion for densification based on per-pixel errors and introduce a mechanism to control the total number of primitives generated per scene. We also correct a bias in the current opacity handling in ADC during cloning. Our approach leads to consistent and systematic improvements over previous methods, particularly in perceptual metrics like LPIPS.

%
%
\bibliographystyle{splncs04}
\bibliography{main}

\clearpage
\appendix

\section{Finer-grained analysis of the MipNeRF360 results}
In this section, we provide quantitative results on the MipNeRF360 dataset broken down into per-scene scores. We report PSNR, SSIM and LPIPS scores averaged over 5 runs with standard deviations. 
In \cref{tab:scene-res}, we compare the performance of our method against standard Gaussian splatting, while in \cref{tab:scene-res-mip} we compare the respective Mip variants. For a fair comparison, we fix the maximum number of primitives for our method to the median number of primitives of the 5 runs of the baseline.

As we can read from both tables, our method outperforms the baselines on all scenes and all metrics, expecting 3 cases: PSNR on flowers and PSNR+SSIM on treehill. In particular, we observe significant gains in terms of LPIPS, which better correlates with the perceptual similarity. Indeed, in the two scenes (flowers and treehill) where PSNR is worse but LPIPS is better than the baseline, our renderings look visually better (see \eg Fig.1 in the main paper). 

\begin{figure}[htb]
    \centering
    \includegraphics[width=\textwidth]{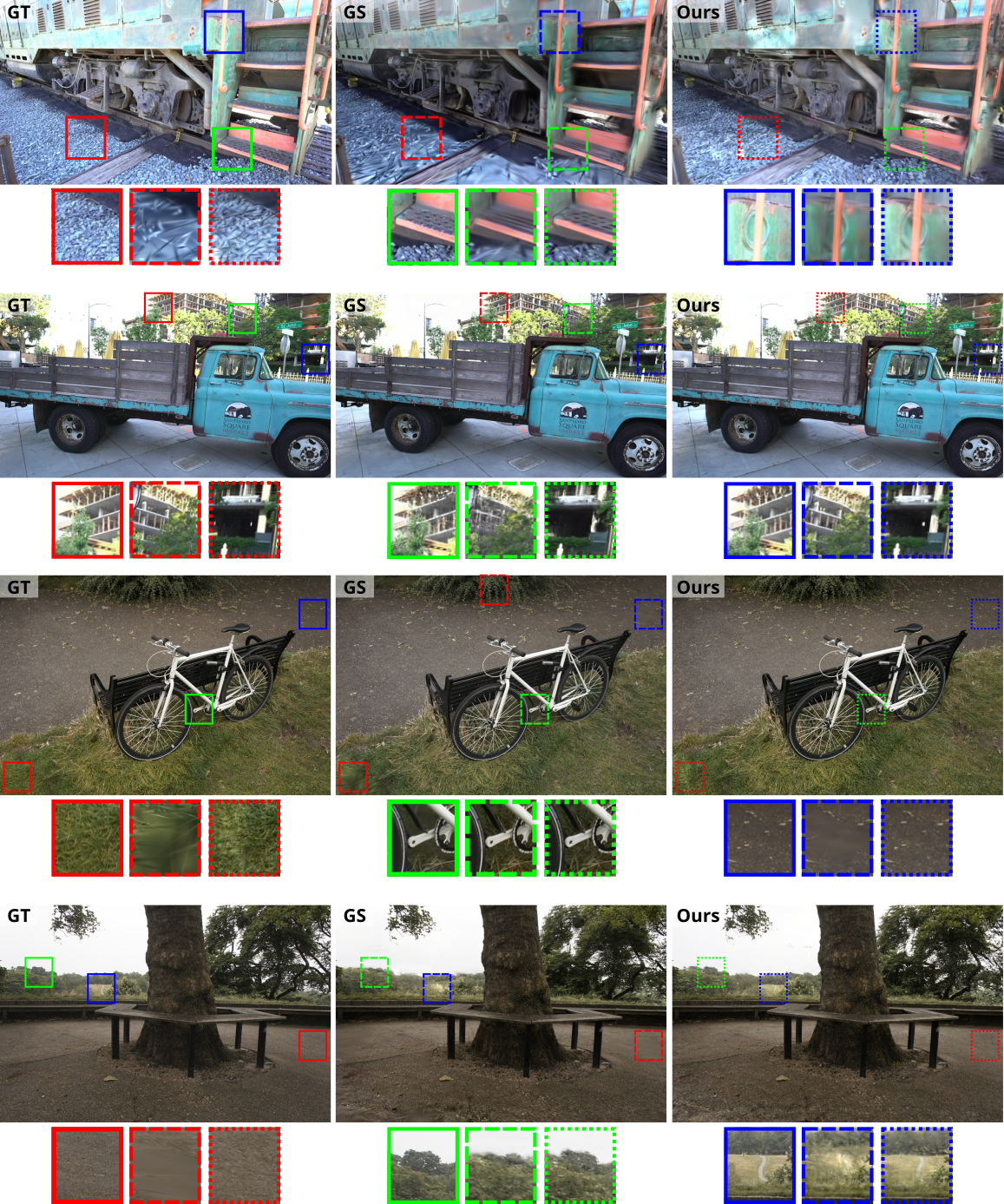}
    \caption{Qualitative results with highlights from Tanks and Temples and MipNeRF360 datasets. We compare ground-truth, standard Gaussian splatting (GS) and our proposed method (Ours).}
    \label{fig:qualitative}
\end{figure}

In \cref{fig:qualitative} we provide additional qualitative results with highligths on some scenes from the Tanks and Temples and MipNeRF360 datasets.

\begin{table}[htb]
    \centering
    \caption{Per-scene quantitative results (PSNR, SSIM, LPIPS) on the MipNeRF360 dataset. We report the results of standard Gaussian Splatting (3DGS) and the proposed method (Ours). Scores are averaged over 5 runs and reported with standard deviations.}
    \label{tab:scene-res}
    \scalebox{.85}{
    \begin{tabular}{l|cc|cc|cc}
    \toprule
    & \multicolumn{2}{c|}{PSNR $\uparrow$} & \multicolumn{2}{c|}{SSIM $\uparrow$} & \multicolumn{2}{c}{LPIPS $\downarrow$}\\
    & 3DGS & Ours & 3DGS & Ours & 3DGS & Ours\\
    \midrule
bicycle&$25.184^{\pm.041}$&${\bf 25.353}^{\pm.025}$&$0.769^{\pm.001}$&${\bf 0.782}^{\pm.001}$&$0.228^{\pm.001}$&${\bf 0.193}^{\pm.001}$\\
bonsai&$32.040^{\pm.205}$&${\bf 32.478}^{\pm.051}$&$0.940^{\pm.001}$&${\bf 0.944}^{\pm.001}$&$0.254^{\pm.001}$&${\bf 0.231}^{\pm.000}$\\
counter&$28.926^{\pm.132}$&${\bf 29.121}^{\pm.091}$&$0.907^{\pm.001}$&${\bf 0.909}^{\pm.001}$&$0.256^{\pm.001}$&${\bf 0.240}^{\pm.001}$\\
flowers&${\bf 21.649}^{\pm.031}$&$21.302^{\pm.035}$&$0.609^{\pm.001}$&${\bf 0.625}^{\pm.001}$&$0.357^{\pm.001}$&${\bf 0.288}^{\pm.001}$\\
garden&$27.332^{\pm.060}$&${\bf 27.622}^{\pm.062}$&$0.867^{\pm.001}$&${\bf 0.872}^{\pm.001}$&$0.120^{\pm.001}$&${\bf 0.109}^{\pm.001}$\\
kitchen&$31.382^{\pm.036}$&${\bf 31.884}^{\pm.103}$&$0.927^{\pm.001}$&${\bf 0.930}^{\pm.000}$&$0.154^{\pm.001}$&${\bf 0.147}^{\pm.000}$\\
room&$31.395^{\pm.107}$&${\bf 31.902}^{\pm.160}$&$0.916^{\pm.001}$&${\bf 0.922}^{\pm.001}$&$0.286^{\pm.001}$&${\bf 0.271}^{\pm.001}$\\
stump&$26.681^{\pm.038}$&${\bf 26.806}^{\pm.030}$&$0.778^{\pm.001}$&${\bf 0.789}^{\pm.001}$&$0.237^{\pm.002}$&${\bf 0.211}^{\pm.001}$\\
treehill&${\bf 22.435}^{\pm.077}$&$22.025^{\pm.042}$&${\bf 0.636}^{\pm.001}$&$0.623^{\pm.002}$&$0.358^{\pm.002}$&${\bf 0.321}^{\pm.001}$\\\midrule
average&$27.447^{\pm.081}$&${\bf 27.611}^{\pm.067}$&$0.817^{\pm.001}$&${\bf 0.822}^{\pm.001}$&$0.250^{\pm.001}$&${\bf 0.223}^{\pm.001}$\\
\bottomrule
    \end{tabular}}
\end{table}

\begin{table}[htb]
    \centering
    \caption{Per-scene quantitative results (PSNR, SSIM, LPIPS) on the MipNeRF360 dataset. We report the results of Mip-splatting (MipS) and our method in its Mip variant (OursMip). Scores are averaged over 5 runs and reported with standard deviations.}
    \label{tab:scene-res-mip}
    \scalebox{.85}{
    \begin{tabular}{l|cc|cc|cc}
    \toprule
    & \multicolumn{2}{c|}{PSNR $\uparrow$} & \multicolumn{2}{c|}{SSIM $\uparrow$} & \multicolumn{2}{c}{LPIPS $\downarrow$}\\
    & MipS & OursMip & MipS & OursMip & MipS & OursMip\\
    \midrule
bicycle&$25.299^{\pm.030}$&${\bf 25.513}^{\pm.048}$&$0.771^{\pm.001}$&${\bf 0.786}^{\pm.001}$&$0.231^{\pm.001}$&${\bf 0.190}^{\pm.001}$\\
bonsai&$32.139^{\pm.103}$&${\bf 32.480}^{\pm.087}$&$0.940^{\pm.001}$&${\bf 0.944}^{\pm.001}$&$0.253^{\pm.001}$&${\bf 0.229}^{\pm.000}$\\
counter&$29.032^{\pm.125}$&${\bf 29.182}^{\pm.048}$&$0.907^{\pm.001}$&${\bf 0.910}^{\pm.000}$&$0.256^{\pm.001}$&${\bf 0.238}^{\pm.000}$\\
flowers&${\bf 21.729}^{\pm.021}$&$21.474^{\pm.049}$&$0.612^{\pm.001}$&${\bf 0.629}^{\pm.001}$&$0.360^{\pm.001}$&${\bf 0.291}^{\pm.001}$\\
garden&$27.430^{\pm.029}$&${\bf 27.698}^{\pm.051}$&$0.870^{\pm.000}$&${\bf 0.873}^{\pm.000}$&$0.120^{\pm.000}$&${\bf 0.110}^{\pm.001}$\\
kitchen&$31.343^{\pm.127}$&${\bf 31.880}^{\pm.111}$&$0.928^{\pm.001}$&${\bf 0.930}^{\pm.001}$&$0.154^{\pm.001}$&${\bf 0.147}^{\pm.001}$\\
room&$31.382^{\pm.129}$&${\bf 32.042}^{\pm.120}$&$0.916^{\pm.003}$&${\bf 0.924}^{\pm.001}$&$0.287^{\pm.002}$&${\bf 0.269}^{\pm.001}$\\
stump&$26.701^{\pm.055}$&${\bf 26.778}^{\pm.033}$&$0.779^{\pm.002}$&${\bf 0.788}^{\pm.001}$&$0.239^{\pm.001}$&${\bf 0.212}^{\pm.000}$\\
treehill&${\bf 22.556}^{\pm.050}$&$22.270^{\pm.041}$&${\bf 0.639}^{\pm.001}$&$0.627^{\pm.001}$&$0.362^{\pm.002}$&${\bf 0.321}^{\pm.001}$\\\midrule
average&$27.512^{\pm.074}$&${\bf 27.702}^{\pm.065}$&$0.818^{\pm.001}$&${\bf 0.823}^{\pm.001}$&$0.251^{\pm.001}$&${\bf 0.223}^{\pm.001}$\\
\bottomrule
    \end{tabular}
    }
\end{table}

\section{Failure of standard Gaussian Splatting even with 10M primitives}
\begin{figure}[htb]
    \centering
    \includegraphics[width=\textwidth]{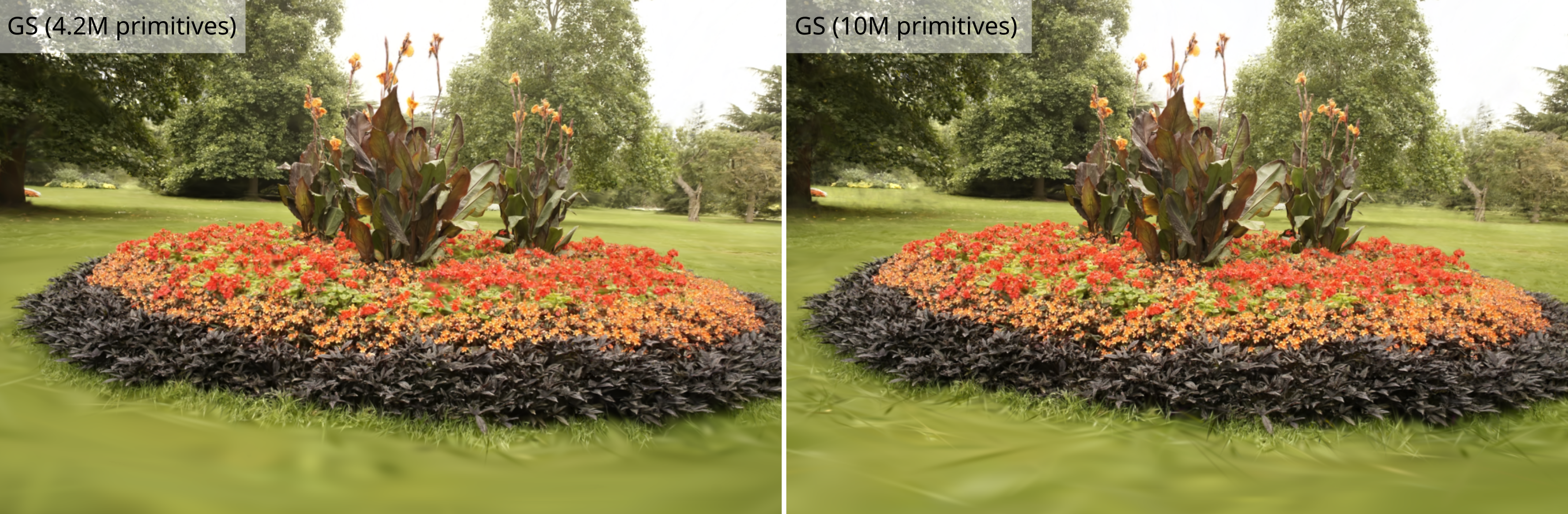}
    \caption{Qualitative result on MipNeRF360-Flowers scene. Left: Gaussian Splatting with standard densification strategy, which yields 4.2M primitives. Right: Gaussian Splatting with our proposed growing strategy and thresholding bypassed to push the number of primitives to 10M.}
    \label{fig:flowers10M}
\end{figure}

We test the hypothesis that the standard Gaussian Splatting (GS) densification logic fails to densify the MipNeRF360-Flowers scene even with 10M primitives. In order to enable GS to reach the desired number of primitives, we bypass the thresholding mechanism and use instead our proposed growing strategy. In \cref{fig:flowers10M}, we report the qualitative results on a particularly difficult validation view for GS. On the left, we see the outcome by using the standard GS algorithm, which yields very blurry grass. The standard approach uses 4.2M primitives for the scene. On the right, we show the result we obtain by pushing the number of primitives to 10M. We observe a slight increase in the number of primitives in the critical area, but the result stays substantially very blurred. This indicates that underrepresented areas might score extremely low if we use the gradient-based densification strategy proposed in GS, to a level that even with 10M primitives we do not reach the point where sufficient densification is triggered.
\section{Ablation of different densification guiding error}
In \cref{tab:guiding-error}, we compare the results of our method and standard Gaussian splatting, when we use $\ell_1$ as the guiding error. We run experiments on the MipNeRF360 dataset. We report the usual metrics averaged over 5 runs and report also standard deviation. Our method still outperforms the baseline on all scenes and all metrics, with a couple of more exceptions compared to using SSIM as the guiding error.

\begin{figure}
    \centering
    \includegraphics[width=.8\textwidth]{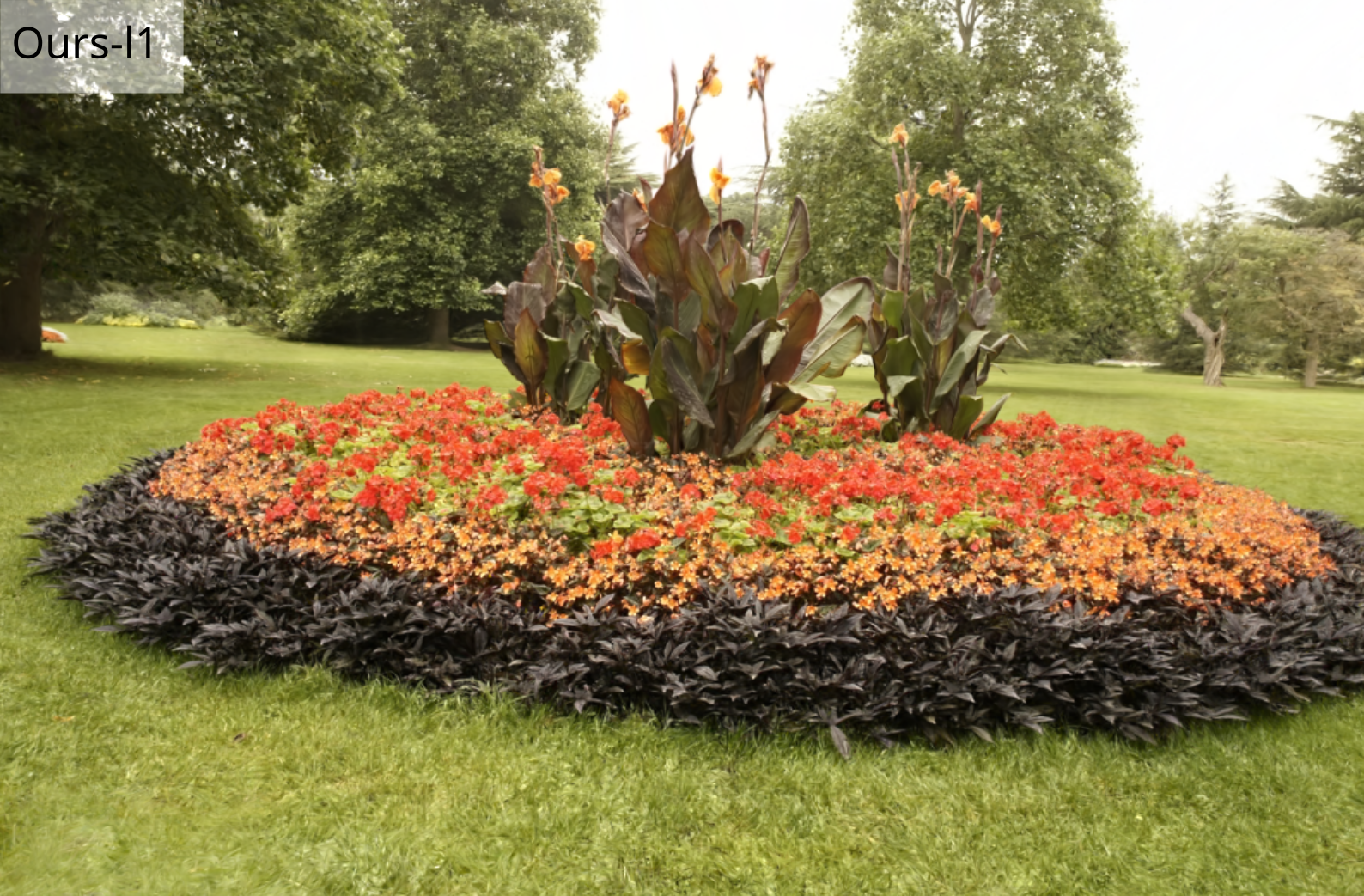}
    \caption{Qualitative result on a validation image from flowers with $\ell_1$ as the densification guiding error.}
    \label{fig:flowers-l1}
\end{figure}

In \cref{fig:flowers-l1}, we also show that despite using an error that does not strongly penalize blurred areas, our method can still reconstruct the grass that is very blurry with standard Gaussian splatting.

\begin{table}[htb]
    \centering
    \caption{Per-scene quantitative results (PSNR, SSIM, LPIPS) on the MipNeRF360 dataset. We report the results of our method with $\ell_1$ as the guiding error (Ours-$\ell_1$) and Gaussian splatting (3DGS). Scores are averaged over 5 runs and reported with standard deviations.}
    \label{tab:guiding-error}
    \scalebox{.85}{
    \begin{tabular}{l|cc|cc|cc}
    \toprule
    & \multicolumn{2}{c|}{PSNR $\uparrow$} & \multicolumn{2}{c|}{SSIM $\uparrow$} & \multicolumn{2}{c}{LPIPS $\downarrow$}\\
    & 3DGS & Ours-$\ell_1$ & 3DGS & Ours-$\ell_1$ & 3DGS & Ours-$\ell_1$\\
    \midrule
bicycle&$25.184^{\pm.041}$&${\bf 25.39}^{\pm.036}$&$0.769^{\pm.001}$&${\bf 0.786}^{\pm.001}$&$0.228^{\pm.001}$&${\bf 0.202}^{\pm.001}$\\
bonsai&$32.040^{\pm.205}$&${\bf 32.22}^{\pm.238}$&$0.940^{\pm.001}$&${\bf 0.941}^{\pm.002}$&$0.254^{\pm.001}$&${\bf 0.244}^{\pm.001}$\\
counter&$28.926^{\pm.132}$&${\bf 29.11}^{\pm.045}$&${\bf 0.907}^{\pm.001}$&$0.906^{\pm.000}$&$0.256^{\pm.001}$&${\bf 0.249}^{\pm.001}$\\
flowers&${\bf 21.649}^{\pm.031}$&$21.37^{\pm.036}$&$0.609^{\pm.001}$&${\bf 0.640}^{\pm.001}$&$0.357^{\pm.001}$&${\bf 0.288}^{\pm.001}$\\
garden&$27.332^{\pm.060}$&${\bf 27.64}^{\pm.031}$&$0.867^{\pm.001}$&${\bf 0.876}^{\pm.001}$&$0.120^{\pm.001}$&${\bf 0.109}^{\pm.001}$\\
kitchen&$31.382^{\pm.036}$&${\bf 31.92}^{\pm.025}$&$0.927^{\pm.001}$&${\bf 0.931}^{\pm.001}$&$0.154^{\pm.001}$&${\bf 0.153}^{\pm.001}$\\
room&$31.395^{\pm.107}$&${\bf 31.43}^{\pm.367}$&$0.916^{\pm.001}$&${\bf 0.919}^{\pm.003}$&${\bf 0.286}^{\pm.001}$&$0.290^{\pm.002}$\\
stump&$26.681^{\pm.038}$&${\bf 26.84}^{\pm.059}$&$0.778^{\pm.001}$&${\bf 0.794}^{\pm.001}$&$0.237^{\pm.002}$&${\bf 0.212}^{\pm.001}$\\
treehill&${\bf 22.435}^{\pm.077}$&$22.07^{\pm.017}$&${\bf 0.636}^{\pm.001}$&$0.621^{\pm.002}$&$0.358^{\pm.002}$&${\bf 0.338}^{\pm.001}$\\\midrule
average&$27.447^{\pm.081}$&${\bf 27.55}^{\pm.095}$&$0.817^{\pm.001}$&${\bf 0.824}^{\pm.001}$&$0.250^{\pm.001}$&${\bf 0.232}^{\pm.001}$\\\bottomrule
    \end{tabular}}
\end{table}

\end{document}